\documentclass[a4paper,11pt]{article}

\usepackage{latexsym}
\usepackage{amsmath}
\usepackage{amsfonts}
\usepackage{amssymb} 
\usepackage{amsthm}
\usepackage{multirow}
\usepackage{url}
\usepackage{subfigure}

\usepackage{lineno}
\usepackage{setspace}

\usepackage{algorithm}
\usepackage{algorithmic}

\usepackage{natbib}
\usepackage{authblk}

\usepackage[svgnames]{xcolor}
\usepackage{tikz}

\usepackage{gb4e}
\usepackage{fullpage}

\usepackage{qtree}
\usepackage{gb4e}

\usepackage{xytree}

\pgfdeclarelayer{edgelayer}
\pgfdeclarelayer{nodelayer}
\pgfsetlayers{edgelayer,nodelayer,main}

\tikzstyle{none}=[inner sep=0pt]
\tikzstyle{plain}=[inner sep=0pt]
\tikzstyle{every picture}=[baseline=(current bounding box).east,scale=0.5,node distance=5mm]

\newcommand{\ov}{\overrightarrow} 

\title{Compositional Operators in Distributional Semantics\footnote{Accepted for publication in \textit{Springer Science Reviews} journal. The final version will be available at \texttt{link. springer.com}.}}

\author{Dimitri Kartsaklis} 
\affil{Department of Computer Science, University of Oxford }
\affil{ \texttt{dimitri.kartsaklis@cs.ox.ac.uk}}

\begin{document}

\newcommand{\tikzfig}[1]{
\InputIfFileExists{#1.tex}{}{\input{./#1.tex}}
}

\newcommand{\ctikzfig}[1]{%
\begin{center}\rm
  
\InputIfFileExists{#1.tex}{}{\input{./#1.tex}}

\end{center}}

\date{}
\maketitle


\begin{abstract}
\noindent
This survey presents in some detail the main advances that have been recently taking place in Computational Linguistics towards the unification of the two prominent semantic paradigms: the compositional formal semantics view and the distributional models of meaning based on vector spaces. After an introduction to these two approaches, I review the most important models that aim to provide compositionality in distributional semantics. Then I proceed and present in more detail a particular framework \citep{Coeckeetal} based on the abstract mathematical setting of category theory, as a more complete example capable to demonstrate the diversity of techniques and scientific disciplines that this kind of research can draw from. This paper concludes with a discussion about important open issues that need to be addressed by the researchers in the future.
\end{abstract}

\begin{center}
 vector space models; formal semantics; category theory; compact closed categories
\end{center}

\section{Introduction}
\label{sec:intro}

\noindent The recent developments on the syntactical and morphological analysis of natural language text constitute the first step towards a more ambitious goal, that of assigning a proper form of \textit{meaning} to arbitrary text compounds. Indeed, for certain really ``intelligent'' applications, such as machine translation, question-answering systems, paraphrase detection, or automatic essay scoring, to name just a few, there will always exist a gap between raw linguistic information (such as part-of-speech labels, for example) and the knowledge of the real world that is needed for the completion of the task in a satisfactory way. Semantic analysis has exactly this role, aiming to close (or reduce as much as possible) this gap by linking the linguistic information with semantic representations that embody this elusive real-world knowledge.

The traditional way of adding semantics to sentences is a syntax-driven compositional approach: every word in the sentence is associated with a primitive symbol or a predicate, and these are combined to larger and larger logical forms based on the syntactical rules of the grammar. At the end of the syntactical analysis, the logical representation of the whole sentence is a complex formula that can be fed to a theorem prover for further processing. Although such an approach seems intuitive, it has been shown that it is rather inefficient for any practical application (for example, \cite{bos2006logical} get very low recall scores for a textual entailment task). Even more importantly, the meaning of the atomic units (words) is captured in an axiomatic way, namely by ad-hoc unexplained primitives that have nothing to say about the real semantic value of the specific words. 

On the other hand, distributional models of meaning work by building co-occurrence vectors for every word in a corpus based on its context, following Firth's intuition that ``you should know a word by the company it keeps'' \citep{Firth}. These models have been proved useful in many natural language tasks (see Section \ref{sec:word2sentence}) and can provide concrete information for the words of a sentence, but they do not scale up to larger constituents of text, such as phrases or sentences. Given the complementary nature of these two distinct approaches, it is not a surprise that compositional abilities of distributional models have been the subject of much discussion and research in recent years. Towards this purpose researchers exploit a wide variety of techniques, ranging from simple mathematical operations like addition and multiplication to neural networks and even category theory. The purpose of this paper is to provide a concise survey of the developments that have been taking place towards the goal of equipping distributional models of meaning with compositional abilities. 

The plan is the following: In Sections \ref{sec:compsem} and \ref{sec:dissem} I provide an introduction to compositional and distributional models of meaning, respectively, explaining the basic principles and assumptions on which they rely. Then I proceed to review the most important methods aiming towards their unification (Section \ref{sec:compdistr}). As a more complete example of such a method (and as a demonstration of the multidisciplinarity of Computational Linguistics), Section \ref{sec:disco} describes the framework of \cite{Coeckeetal}, based on the abstract setting of category theory. Section \ref{sec:sentencespace} provides a closer look to the form of a sentence space, and how our sentence-producing functions (i.e. the verbs) can be built from a large corpus. Finally, Section \ref{sec:challenges} discusses important philosophical and practical open questions and issues that form part of the current and future research.

\section{Compositional semantics}
\label{sec:compsem}

Compositionality in semantics offers an elegant way to address the inherent property of natural language to produce infinite structures (phrases and sentences) from finite resources (words). The \textit{principle of compositionality} states that the meaning of a complex expression can be determined by the meanings of its constituents and the rules used for combining them. This idea is quite old, and glimpses of it can be spotted even in works of Plato. In his dialogue \textit{Sophist}, Plato argues that a sentence consists of a noun and a verb, and that the sentence is true if the verb denotes the action that the noun is currently performing. In other words, Plato argues that (a) a sentence has a structure; (b) the parts of the sentence have different functions; (c) the meaning of the sentence is determined by the function of its parts. Nowadays, this intuitive idea is often attributed to Gottlob Frege, who expresses similar views in his ``Foundations of Mathematics'', originally published in 1884. In an undated letter to Philip Jourdain, included in ``Philosophical and Mathematical Correspondence'' (\citeyear{fregeletter}), Frege provides an explanation for the reason this idea seems so intuitive:

\begin{quote}
``The possibility of our understanding propositions which we have never heard before rests evidently on this, that we can construct the sense of a proposition out of parts that correspond to words.''
\end{quote}

This forms the basis of the \textit{productivity} argument, often used as a proof for the validity of the principle: humans only know the meaning of words, and the rules to combine them in larger constructs; yet, being equipped with this knowledge, we are able to produce new sentences that we have never uttered or heard before. Indeed, this task seems natural even for a 3-years old child---however, its formalization in a way reproducible by a computer has been proven nothing but trivial. The modern compositional models owe a lot to the seminal work of Richard Montague (1930-1971), who has managed to present a systematic way of processing fragments of the English language in order to get semantic representations capturing their ``meaning'' \citep{Mon1,Mon2,Mon3}. In his ``Universal Grammar'' (\citeyear{Mon2}), Montague states:

\begin{quote}
``There is in my opinion no important theoretical difference between natural languages and the artificial languages of logicians.''
\end{quote}

Montague supports this claim by detailing a systematization of the natural language, an approach which became known as Montague grammar. To use Montague's method, one would need two things: first, a resource which will provide the logical forms of each specific word (a lexicon); and second, a way to determine the correct order in which the elements in the sentence should be combined in order to end up with a valid semantic representation. A natural way to address the latter, and one traditionally used in computational linguistics, is to use the syntactic structure as a means of driving the semantic derivation (an approach called \textit{syntax-driven semantic analysis}). In other words, we assume that there exists a mapping from syntactic to semantic types, and that the composition in the syntax level implies a similar composition in the semantic level. This is known as the \textit{rule-to-rule hypothesis} \citep{Bach:76}. 

In order to provide an example, I will use the sentence `Every man walks'. We begin from the lexicon, the job of which is to provide a grammar type and a logical form to every word in the sentence:

\begin{exe}
\ex 
\begin{xlist}
\ex\label{every} every $\vdash Det : \lambda P.\lambda Q.\forall x[P(x) \to Q(x)]$
\ex\label{man} man $\vdash N : \lambda y.man(y)$
\ex\label{walks} walks $\vdash  Verb_{IN} : \lambda z.walks(z)$
\end{xlist}
\end{exe}

The above use of formal logic (especially higher-order) in conjunction with $\lambda$-calculus was first introduced by Montague, and from then on it constitutes the standard way of providing logical forms to compositional models. In the above lexicon, predicates of the form $man(y)$ and $walks(z)$ are true if the individuals denoted by $y$ and $z$ carry the property (or, respectively, perform the action) indicated by the predicate. From an extensional perspective, the semantic value of a predicate can be seen as the set of all individuals that carry a specific property: $walks(john)$ will be true if the individual $john$ belongs to the set of all individuals who perform the action of walking. Furthermore, $\lambda$-terms like $\lambda x$ or $\lambda Q$ have the role of placeholders that remain to be filled. The logical form $\lambda y.man(y)$, for example, reflects the fact that the entity which is going to be tested for the property of manhood is still unknown and it will be later specified based on the syntactic combinatorics. Finally, the form in (\ref{every}) reflects the traditional way for representing a universal quantifier in natural language, where the still unknown part is actually the predicates acting over a range of  entities.


In $\lambda$-calculus, function application is achieved via the process of $\beta$-reduction: given two logical forms $\lambda x.t$ and $s$, the application of the former to the latter will produce a version of $t$ where all the free occurrences of $x$ in $t$ have been replaced by $s$. More formally:

\begin{equation}
  (\lambda x.t)s \to_\beta t[x:=s]
\end{equation}

Let us see how we can apply the principle of compositionality to get a logical form for the above example sentence, by repeatedly applying $\beta$-reduction between the semantic forms of text constituents following the grammar rules. The parse tree in (\ref{cfgtree}) below provides us a syntactic analysis:

\singlespace
\qtreecenterfalse
\begin{exe}
\ex\label{cfgtree} \Tree [ .\textit{S} [ .\textit{NP} [ .\textit{Det} Every ] [.\textit{N} man ] ] [.\textit{Verb}_{IN} walks ] ] 
\end{exe}

Our simple context-free grammar (CFG) consists of two rules:

\begin{exe}
  \ex 
  \begin{tabular}{ll}
    $\textit{NP} \to \textit{Det}~N$ & a noun phrase consists of a determiner and a noun\\
      $S \to \textit{NP}~\textit{Verb}_{IN}$ & a sentence consists of a noun phrase and an intransitive verb
      \end{tabular}
\end{exe}

These rules will essentially drive the semantic derivation. Interpreted from a semantic perspective, the first rule states that the logical form of a noun phrase is derived by applying the logical form of a determiner to the logical form of a noun. In other words, $P$ in (\ref{every}) will be substituted by the logical form for \textit{man} (\ref{man}). The details of this reduction are presented below:

\begin{exe}
  \ex\label{everyman}
  \begin{tabular}{ll}
    $\lambda P.\lambda Q.\forall x[P(x) \to Q(x)](\lambda y.man(y))$ & \\
    $ ~~\to_\beta \lambda Q.\forall x[(\lambda y.man(y))(x) \to Q(x)]$ & $P:= \lambda y.man(y)$\\
    $ ~~\to_\beta \lambda Q.\forall x[man(x) \to Q(x)]$ & $y := x$
  \end{tabular}  
\end{exe}
 
Similarly, the second rule signifies that the logical form of the whole sentence is derived by the combination of the logical form of the noun phrase as computed in (\ref{everyman}) above with the logical form of the intransitive verb (\ref{walks}):

\begin{exe}
  \ex\label{everymanwalks}
  \begin{tabular}{ll}
    $\lambda Q.\forall x[man(x) \to Q(x)](\lambda z.walks(z))$ & \\
    $~~\to_\beta \forall x[man(x) \to (\lambda z.walks(z))(x)]$ & $Q:= \lambda z.walks(z)$ \\
    $~~\to_\beta \forall x[man(x) \to walks(x)]$ & $z := x$ 
  \end{tabular}
\end{exe}

Thus we have arrived at a logical form which can be seen as a semantic representation of the whole sentence. The tree below provides a concise picture of the complete semantic derivation:

\singlespace
\qtreecenterfalse
\begin{exe}
\ex\label{cfgtreesem} \Tree [ .{\textit{S}\\$\forall x[man(x) \to walks(x)]$} [ .{\textit{NP}\\$\lambda Q.\forall x[man(x) \to Q(x)]$} [ .{\textit{Det}\\$\lambda P.\lambda Q.\forall x[P(x) \to Q(x)]$} Every ] [.{\textit{N}\\$\lambda y.man(y)$} man ] ] [.{\textit{Verb}_{IN}\\$\lambda z.walks(z)$} walks ] ] 
\end{exe}

A logical form such as $\forall x[man(x) \to walks(x)]$ simply states the truth (or falseness) of the expression given the sub-expressions and the rules for composing them. It does not provide any quantitative interpretation of the result (e.g. grades of truth) and, even more importantly, leaves the meaning of words as unexplained primitives ($man$, $walks$ etc). In the next section we will see how distributional semantics can fill this gap.

\section{Distributional semantics}
\label{sec:dissem}

\subsection{The distributional hypothesis}
\label{sec:dishypothesis}

The distributional paradigm is based on the \textit{distributional hypothesis} \citep{Harris}, stating that words that occur in the same context have similar meanings. Various forms of this popular idea keep recurring in the literature: \cite{Firth} calls it collocation, while Frege himself states that ``never ask for the meaning of a word in isolation, but only in the context of a proposition.'' \citep{frege1980foundations}. The attraction of this principle in the context of Computational Linguistics is that it provides a way of concretely representing the meaning of a word via mathematics: each word is a vector whose elements show how many times this word occurred in some corpus at the same context with every other word in the vocabulary. If, for example, our basis is $\{cute,sleep,finance,milk\}$, the vector for word `cat' could have the form $(15,7,0,22)$ meaning that `cat' appeared 15 times together with `cute', 7 times with `sleep' and so on. More formally, given an orthonormal basis $\{\ov{n_i}\}_i$ for our vector space, a word is represented as:

\begin{equation}
  \ov{word} = \sum_i c_i\ov{n_i}
\end{equation}

\noindent
where $c_i$ is the coefficient for the $i$th basis vector. As mentioned above, in their simplest form these coefficients can be just co-occurrence counts, although in practice a function on counts is often used in order to remove some of the unavoidable frequency bias. A well-known measure is the information-theoretic \textit{point-wise mutual information} (PMI), which can reflect the relationship between a context word $c$ and a target word $t$ as follows:

\begin{equation}
  \text{PMI}(c,t) = \log\frac{p(c|t)}{p(t)}
\end{equation}

On the contrary with compositional semantics which leaves the meaning of lexical items unexplained, a vector like the one used above for `cat' provides some concrete information about the meaning of the specific word: cats are cute, sleep a lot, they really like milk, and they have nothing to do with finance. Additionally, this quantitative representation allows us to compare the meanings of two words, e.g. by computing the cosine distance of their vectors, and evaluate their semantic similarity. The cosine distance is a popular choice for this task (but not the only one) and is given by the following formula:

\begin{equation}
  \text{sim}(\ov{v},\ov{u}) = \cos({\ov{v},\ov{u}}) = \frac{ \langle \ov{v} | \ov{u} \rangle}{\Vert\ov{v}\Vert \Vert\ov{u}\Vert}
\end{equation}

\noindent
where $\langle \ov{v} | \ov{u} \rangle$ denotes the dot product between $\ov{v} and \ov{u}$, and $\Vert \ov{v} \Vert$ the magnitude of $\ov{v}$. 

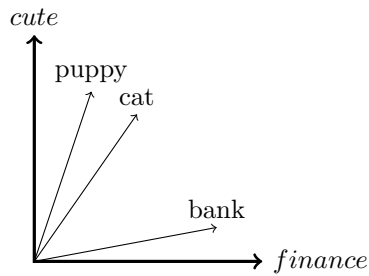
\begin{figure}[h]
 \centering
 \small
 \begin{tikzpicture}[scale=3.0]
    \draw [<->,very thick] (0,2) node (yaxis) [above] {$cute$}
        |- (2,0) node (xaxis) [right] {$finance$};
    \draw [->] (0,0) -- (0.5,1.5) node (puppy) [above] {puppy};
    \draw [->] (0,0) -- (0.9,1.3) node (cat) [above] {cat};
    
    \draw [->] (0,0) -- (1.6,0.3) node (bank) [above] {bank};
 \end{tikzpicture}
 \normalsize
\caption{A toy ``vector space'' for demonstrating semantic similarity between words.}
\label{fig:sim}
\end{figure}

As an example, in the 2-dimensional vector space of Figure \ref{fig:sim} below we see that `cat' and `puppy' are close together (and both of them closer to the basis `cute'), while `bank' is closer to basis vector `finance'.

\begin{figure}[h!]
 \centering
 \includegraphics[scale=0.6,trim = 5.7cm 7cm 8cm 3.5cm, clip]{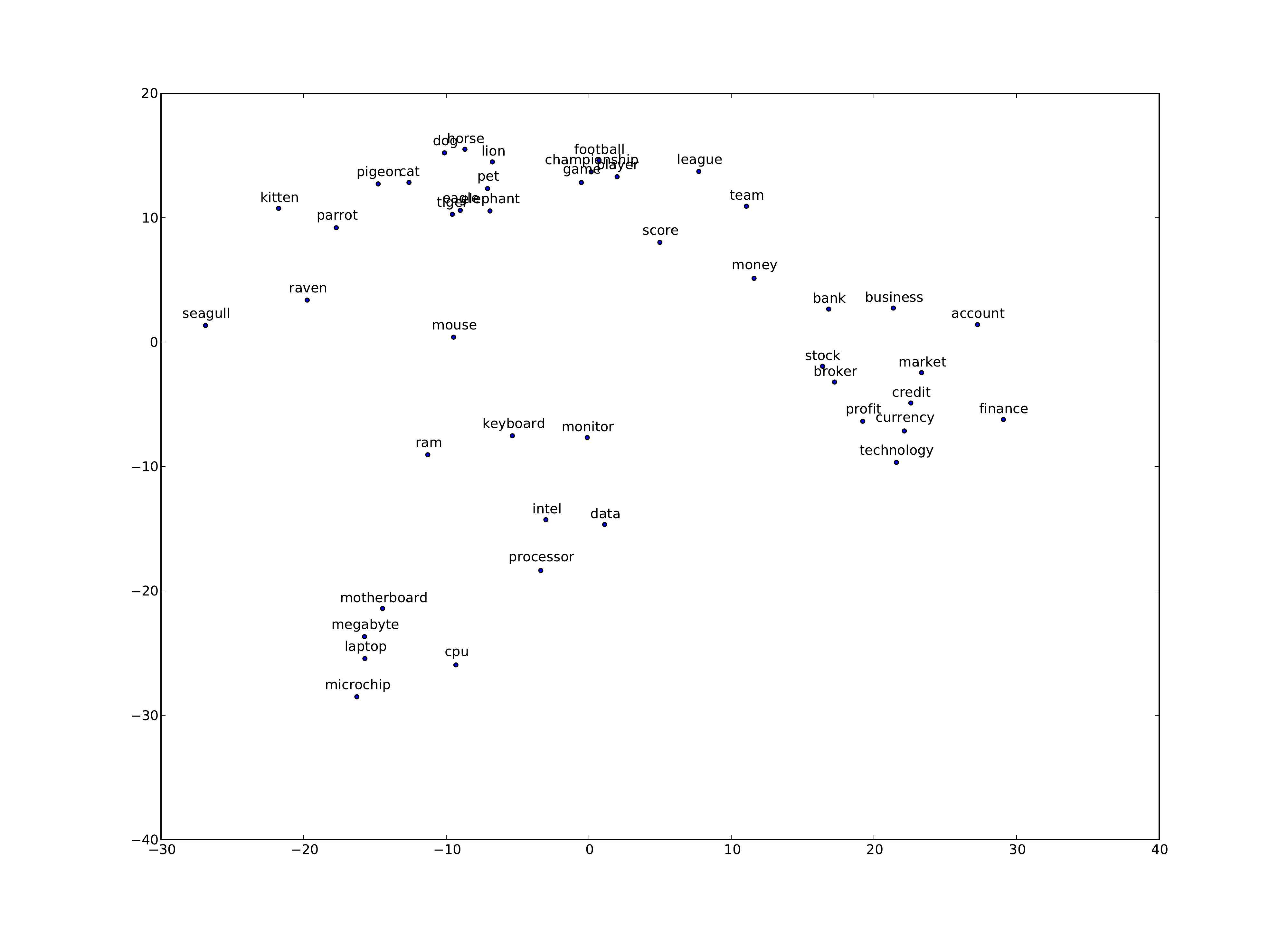}
 \caption{Visualization of a word space in two dimensions (original vectors are 2000-dimensional vectors created from BNC).}
 \label{fig:proj}
\end{figure}

A more realistic example is shown in Figure \ref{fig:proj}. The points in this space represent real distributional word vectors created from British National Corpus (BNC)\footnote{BNC is a 100 million-word text corpus consisting of samples of written and spoken English. It can be found online at \texttt{http://www.natcorp.ox.ac.uk/}.}, originally 2,000-dimensional and projected onto two dimensions for visualization. Note how words form distinct groups of points according to their semantic correlation. Furthermore, it is interesting to see how ambiguous words behave in these models: the ambiguous word `mouse' (with the two meanings to be that of a rodent and of a computer pointing device), for example, is placed almost equidistantly from the group related to IT concepts (lower left part of the diagram) and the animal group (top left part of the diagram), having a meaning that can be indeed seen as the average of both senses.

\subsection{Forms of word spaces}
\label{sec:vectorspaces}

In the simplest form of a word space (a vector space for words), the context of a word is a set containing all the words that occur within a certain distance from the target word, for example within a 5-word window. Models like these have been extensively studied and implemented the past years, see for example the work of \cite{lowe2001towards} and \cite{lowe2000direct}. However, although such an approach is simple, intuitive and computationally efficient, it is not optimal, since it assumes that every word within this window will be semantically relevant to the target word, while treats every word outside of the window as irrelevant. Unfortunately, this is not always the case. Consider for example the following phrase: 

\begin{exe}
\ex The movie I saw and John said he really likes
\end{exe}

Here, a word-based model can not efficiently capture the long-ranged dependency between `movie' and `likes'; on the other hand, since `said' is closer to `movie' it is more likely to be considered as part of its context, despite the fact that the semantic relationship between the two words is actually weak. Cases like the above suggest that a better approach for the construction of the model would be to take into account not just the surface form of context words, but also the specific grammatical relations that hold between them and the target word. A model like this, for example, will be aware that `movie' is the object of `likes', so it could safely consider the latter as part of the context for the former, and vice versa. This kind of observations motivated many researchers to experiment with vector spaces based not solely on the surface forms of words but also on various morphological and syntactical properties of the text.

One of the earliest attempts to add syntactical information in a word space was that of \cite{grefenstette1994}, who used a structured vector space with a basis constructed by grammatical properties such as `subject-of-buy' or `argument-of-useful', denoting that the target word has occurred in the corpus as subject of the verb `buy' or as argument of the adjective `useful'. The weights of a word vector were binary, either 1 for at least one occurrence or 0 otherwise. \cite{lin1998} moves one step further, replacing the binary weights with frequency counts. Following a different path, \cite{ErkPado} argue that a single vector is not enough to catch the meaning of a word; instead, the vector of a word is accompanied by a set of vectors $R$ representing the lexical preferences of the word for its arguments positions, and a set of vectors $R^{-1}$, denoting the inverse relationship, that is, the usage of the word as argument in the lexical preferences of other words. Subsequent works by \cite{thater2010,thater2011word} present a version of this idea extended with the inclusion of grammatical dependency contexts.

In an attempt to provide a generic distributional framework, \cite{pado2007} presented a model based on dependency relations between words. The interesting part of this work is that, given the proper parametrization, it is indeed able to essentially subsume a large amount of other works on the same subject. For this reason, it is worth of a more detailed description which I provide in the next section.

\subsection{A dependency-based model}
\label{sec:dependency}

The generic framework of \cite{pado2007} treats each sentence as a directed graph, the nodes of which are the words and the edges the dependency relations that hold between them. In (\ref{deptree}) below we can see a typical dependency diagram; each arrow starts from a head word and ends on a dependent, whereas the labels denote the specific relationships that hold between the words:

\begin{exe}
\ex\label{deptree} \xytext{
  \xybarnode{Peter} &~~~&
  \xybarnode{and}
    \xybarconnect(UL,U){-2}"_{\small conj}"
    \xybarconnect(UR,U){2}"^{\small conj}"
    &~~~&
  \xybarnode{Mary} &~~~&
  \xybarnode{bought}
    \xybarconnect[8](UL,U){-4}"_{\small subj}"
    \xybarconnect[13]{6}"^{\small punct}"
    \xybarconnect[8](UR,U){4}"^{\small obj}"
    &~~~&
  \xybarnode{a} &~~~&
  \xybarnode{car}
    \xybarconnect(UL,U){-2}"_{\small det}"
    &~~~&
  \xybarnode{.}
}
\end{exe}

The context of a target word consists of all dependency paths that start from this specific word and end to some other word within a given distance. A valid path should not be cyclic, in order to avoid linguistically meaningless situations such as paths of infinite length or paths consisting of unconnected fragments. Let us see an example for the simple sentence `dogs chase cats', whose dependency relations set is $\{object(chase,cats)$, $subject(chase,dogs)\}$. The set of paths for the target word `dogs', then, would be $\{dogs-chase,dogs-chase-cats\}$ (note that the direction information is dropped at this point---the paths are treated as \textit{undirected} in order to catch relationships between e.g. a subject and an object, which otherwise would be impossible). The creation of a vector space involves the following steps:

\begin{enumerate}

\item For each target word $t$, collect all the undirected paths that start from this word. This will be the initial context for $t$, denoted by $\Pi_t$.

\item Apply a context selection function $cont:W \to 2^{\Pi_t}$ (where $W$ is the set of all tokens of type $t$), which assigns to $t$ a subset of the initial context. Given a word-window $k$, for example, this function might be based on the absolute difference between the position of $t$ and the position of the end word for each path.

\item For every path $\pi \in cont(t)$, specify a relative importance value by applying a path value function of the form $v:\Pi \to \mathbb{R}$.

\item For every $\pi \in cont(t)$, apply a basis-mapping function $\mu: \Pi \to B$, which maps each path $\pi$ to a specific basis element.

\item Calculate the co-occurrence frequency of $t$ with a basis element $b$ by a function $f:B \times T \to \mathbb{R}$, defined as:

\begin{equation}
  f(b,t)  = \sum\limits_{w \in W(t)}~\sum\limits_{\pi \in cont(w) \wedge \mu(\pi)=b} v(\pi)
\end{equation}

\item Finally, and in order to remove potential frequency bias due to raw counts, calculate the log-likelihood ratio G$^2$ \citep{dunning1993} for all basis elements.

\end{enumerate}

An appealing future of this framework is that it is fully parametrized by using different forms of the functions $cont$, $v$, and $\mu$. In order to avoid sparsity problems, the authors chose to use for their experiments a fixed basis mapping function that maps every dependency path to its ending word. This results in a vector space with words as basis elements, quite similar to the word-based setting described in Section \ref{sec:vectorspaces}. There is, however, the important difference that the inclusion criterion of a word into the context of another word is not simply the co-occurrence of the two words within a given window, but the fact that they are related through some dependency path. The authors experimented with various similarity measures and different forms of the functions $cont$ and $v$, producing similarity scores for the well-known benchmark dataset of \cite{rubenstein1965}. This dataset consists of 65 pairs of nouns; the task is to estimate the similarity between the two nouns of each pair, and compare the results with a gold reference prepared by human annotators. Their best results (expressed as the Pearson correlation of model scores with human scores) came from the information-theoretic similarity measure of \cite{lin1998}, a context selection function that retains paths with length $\le 3$, and a simple path value function that assigns 1 to paths of length 1 (since these paths correspond to the most direct and strong dependencies), and fractions to longer paths. More formally, the best-performing model has the following form (including the selected basis mapping function):

\begin{equation}
  cont(t) = \{ \pi \in \Pi_t |~\Vert\pi\Vert \le 3 \}~~,~~
  v(\pi) = \frac{1}{\Vert \pi \Vert}~~,~~
  \mu(\pi) = end(\pi)
 \label{equ:depsuccess}
\end{equation}

\noindent where $end(\pi)$ denotes the ending word of path $\pi$. 

\subsection{From words to sentence}
\label{sec:word2sentence}

Distributional models of meaning have been widely studied and successfully applied on a variety of language tasks, especially during the last decade with the availability of large-scale corpora, like Gigaword \citep{gigaword} and ukWaC \citep{ukwac}, which provide a reliable resource for training the vector spaces. For example, \cite{Landauer} use vector space models in order to model and reason about human learning rates in language; \cite{Schutze} performs word sense induction and disambiguation; \cite{Curran} shows how distributional models can be applied to automatic thesaurus extraction; \cite{Manning} discuss possible applications in the context of information retrieval.

However, due to the infinite capacity of a natural language to produce new sentences from finite resources, no corpus, regardless its size, can be used for providing vector representations to anything but very small text fragments, usually only to words. Under this light, the provision of distributional models with compositional abilities similar to what was described in Section \ref{sec:compsem} seems a very appealing solution that could offer the best of both worlds in a unified manner. The goal of such a system would be to use the compositional rules of a grammar, as described in Section \ref{sec:compsem}, in order to combine the context vectors of the words to vectors of larger and larger text constituents, up to the level of a sentence. A sentence vector, then, could be compared with other sentence vectors, providing a way for assessing the semantic similarity between sentences as if they were words. The benefits of such a feature are obvious for many natural language processing tasks, such as paraphrase detection, machine translation, information retrieval, and so on, and in the following section I am going to review all the important approaches and current research towards this challenging goal.

\section{Compositionality in distributional approaches}
\label{sec:compdistr}

\subsection{Vector mixture models}
\label{sec:vecmix}

The transition from word vectors to sentence vectors implies the existence of a composition operation that can be applied between text constituents according to the rules of the grammar: the composition of `red' and `car' into the adjective-noun compound `red car', for example, should produce a new vector derived from the composition of the context vectors for `red' and `car'. Since we work with vector spaces, the candidates that first come to mind is vector addition and vector (point-wise) multiplication. Indeed, \cite{Lapata} present and test various models, where the composition of vectors is based on these two simple operations. Given two word vectors $\ov{w_1}$ and $\ov{w_2}$ and assuming and orthonormal basis $\{\ov{n_i}\}_i$, the multiplicative model computes the meaning vector of the new compound as follows:

\begin{equation}
\label{equ:mult}
   \ov{w_1w_2} = \ov{w_1} \odot \ov{w_2} = \sum\limits_i c_i^{w_1}c_i^{w_2}\ov{n_i} 
\end{equation}

\begin{figure}[h]
  \centering
  \includegraphics[scale=0.6,trim=0cm 0cm 10cm 1.3cm,clip]{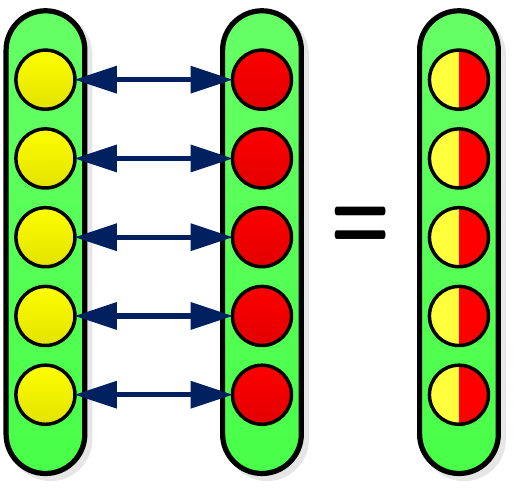}
  \caption{Vector mixture models. Each element in the output vector is a mixture of the corresponding elements in the input vectors.}
  \label{fig:vecmix}
\end{figure}

\noindent
Similarly, the meaning according to the additive model is:

\begin{equation}
\label{equ:add}
  \ov{w_1w_2} = \alpha \ov{w_1} + \beta \ov{w_2} = \sum\limits_i (\alpha c_i^{w_1}+\beta c_i^{w_2})\ov{n_i}
\end{equation}

\noindent where $\alpha$ and $\beta$ are optional weights denoting the relative importance of each word. The main characteristic of these models is that all word vectors live into the same space, which means that in general there is no way to distinguish between the type-logical identities of the different words. This fact, in conjunction with the element-wise nature of the operators, makes the output vector a kind of \textit{mixture} of the input vectors. Figure \ref{fig:vecmix} demonstrates this; each element of the output vector can be seen as an ``average'' of the two corresponding elements in the input vectors. In the additive case, the components of the result are simply the cumulative scores of the input components. So in a sense the output element embraces both input elements, resembling a union of the input features. On the other hand, the multiplicative version is closer to intersection: a zero element in one of the input vector will eliminate the corresponding feature in the output, no matter how high the other component was.

Vector mixture models constitute the simplest compositional method in distributional semantics. Despite their simplicity, though, (or because of it) these approaches have been proved very popular and useful in many NLP tasks, and they are considered hard-to-beat baselines for many of the more sophisticated models we are going to discuss next. In fact, the comparative study of \cite{blacoe2012} suggests something really surprising: that, for certain tasks, additive and multiplicative models can be almost as much effective as state-of-the-art deep learning models, which will be the subject of Section \ref{sec:deeplearning}. 

\subsection{Tensor product and circular convolution}
\label{sec:tensorproduct}

The low complexity of vector mixtures comes with a price, since the produced composite representations disregard grammar in many different ways. For example, an obvious problem with these approaches is the commutativity of the operators: the models treat a sentence as a ``bag of words'' where the word order does not matter, equating for example the meaning of sentence `dog bites man' with that of `man bites dog'. This fact motivated researchers to seek solutions on non-commutative operations, such as the tensor product between vector spaces. Following this suggestion, which was originated by \cite{Smolensky}, the composition of two words is achieved by a structural mixing of the basis vectors that results in an increase in dimensionality:

\begin{equation}
  \ov{w_1} \otimes \ov{w_2} = \sum\limits_{i,j} c^{w_1}_ic^{w_2}_j(\ov{n_i} \otimes \ov{n_j})
\end{equation}

\cite{ClarkPulman} take this original idea further and propose a concrete model in which the meaning of a word is represented as the tensor product of the word's context vector with another vector that denotes the grammatical role of the word and comes from a different abstract vector space of grammatical relationships. As an example, the meaning of the sentence `dog bites man' is given as:

\begin{equation}
  \ov{dog~bites~man} = 
  (\ov{dog} \otimes \ov{subj}) \otimes \ov{bites} \otimes
  (\ov{man} \otimes \ov{obj})
  \label{equ:clarkpulman}
\end{equation}

Although tensor product models solve the bag-of-words problem, unfortunately introduce a new very important issue: given that the cardinality of the vector space is $d$, the space complexity grows exponentially as more constituents are composed together. With $d=300$, and assuming a typical floating-point machine representation (8 bytes per number), the vector of Equation \ref{equ:clarkpulman} would require $300^5 \times 8 = 1.944 \times 10^{13}$ bytes ($\approx$ 19.5 terabytes). Even more importantly, the use of tensor product as above only allows the comparison of sentences that share the same structure, i.e. there is no way for example to compare a transitive sentence with an intransitive one, a fact that severely limits the applicability of such models.

Using a concept from signal processing, \cite{Plate} suggest the replacement of tensor product by circular convolution. This operation carries the appealing property that its application on two vectors results in a vector of the same dimensions as the operands. Let $\ov{v}$ and $\ov{u}$ be vectors of $d$ elements, the circular convolution of them will result in a vector $\ov{c}$ of the following form:

\begin{equation}
  \ov{c} = \ov{v} \circledast \ov{u} = \sum\limits_{i=0}^{d-1} \left(\sum\limits_{j=0}^{d-1} v_ju_{i-j}\right) \ov{n}_{i+1}
\end{equation}

\noindent where $\ov{n_i}$ represents a basis vector, and the subscripts are modulo-$d$, giving to the operation its circular nature. This can be seen as a compressed outer product of the two vectors. However, a successful application of this technique poses some restrictions. For example, it requires a different interpretation of the underlying vector space, in which ``micro-features'', as PMI weights for each context word, are replaced by what Plate calls ``macro-features'', i.e. features that are represented by whole vectors drawn from a normal distribution. Furthermore, circular convolution is commutative, re-introducing the bag-of-words problem\footnote{Actually, Plate proposes a workaround for the commutativity problem; however this is not quite satisfactory and not specific to his model, since it can be used with any other commutative operation such as vector addition or point-wise multiplication.}. 

\subsection{Tensor-based models}
\label{sec:tensorbased}

\begin{figure}[h]
  \centering
  \includegraphics[scale=0.6,trim=6.5cm 0cm 0cm 1.3cm,clip]{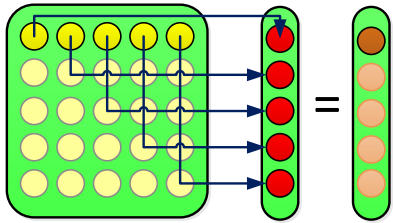}
  \caption{Tensor-based models. The $i$th element in the output vector is computed as the linear combination of the input vector with the $i$th row of the matrix representing the linear map.}
  \label{fig:tensorbased}
\end{figure}

A weakness of vector mixture models and the generic tensor product approach is the symmetric way in which they treat all words, ignoring their special roles in the sentence. An adjective, for example, lives in the same space with the noun it modifies, and both will contribute equally to the output vector representing the adjective-noun compound. However, a treatment like this seems unintuitive; we tend to see relational words, such as verbs or adjectives, as functions acting on a number of arguments rather than entities of the same order as them. Following this idea, a recent line of research represents words with special meaning as linear maps (tensors\footnote{Here, the word \textit{tensor} refers to a geometric object that can be seen as a generalization of a vector in higher dimensions. A matrix, for example, is an order-2 tensor.} of higher order) that apply on one or more arguments (vectors or tensors of lower order). An adjective, for example, is not any more a simple vector but a matrix (a tensor of order 2) that, when matrix-multiplied with the vector of a noun, will return a modified version of it (Figure \ref{fig:tensorbased}).

This approach is based on the well-known \textit{Choi-Jamio{\l}kowsky isomorphism}: every linear map from $V$ to $W$ (where $V$ and $W$ are finite-dimensional Hilbert spaces) stands in one-to-one correspondence with a tensor living in the tensor product space $V \otimes W$. For the case of a multilinear map (a function with more than one arguments), this can be generalized to the following:

\begin{equation}
  f: V_1 \to \dots \to V_j \to V_k \cong V_1 \otimes \dots \otimes V_j \otimes V_k
\end{equation}

In general, the order of the tensor is equal to the number of arguments plus one order for carrying the result; so a unary function (such as an adjective) is a tensor of order 2, while a binary function (e.g. a transitive verb) is a tensor of order 3. The composition operation is based on the inner product and is nothing more than a generalization of matrix multiplication in higher dimensions, a process known as \textit{tensor contraction}. Given two tensors of orders $m$ and $n$, the tensor contraction operation always produces a new tensor of order $n+m-2$. Under this setting, the meaning of a simple transitive sentence can be calculated as follows:

\begin{equation}
  \label{equ:trans}
  \overline{subj~verb~obj} = \ov{subj}^{\mathsf{T}} \times \overline{verb} \times \ov{obj}
\end{equation}

\noindent where the symbol $\times$ denotes tensor contraction. Given that $\ov{subj}$ and $\ov{obj}$ live in $N$ and $\overline{verb}$ lives in $N \otimes S \otimes N$, the above operation will result in a tensor in $S$, which represents the sentence space of our choice (for a discussion about $S$ see Section \ref{sec:sentencespace}). 

Tensor-based models provide an elegant solution to the problems of vector mixtures: they are not bag-of-words approaches and they respect the type-logical identities of special words, following an approach very much aligned with the formal semantics perspective. Furthermore, they do not suffer from the space complexity problems of models based on raw tensor product operations (Section \ref{sec:tensorproduct}), since the tensor contraction operation guarantees that every sentence will eventually live in our sentence space $S$. On the other hand, the highly linguistic perspective they adopt has also a downside: in order for a tensor-based model to be fully effective, an appropriate mapping to vector spaces should has been devised for every functional word, such as prepositions, relative pronouns or logical connectives. As we will see in Sections \ref{sec:functional} and \ref{sec:logical}, this problem is far from trivial; actually it constitutes one of the most important open issues, and at the moment restricts the application of these models on well-defined text structures (for example, simple transitive sentences of the form `subject-verb-object' or adjective-noun compounds).

The notion of a framework where relational words act as linear maps on noun vectors has been formalized by \cite{Coeckeetal} in the abstract setting of category theory and compact closed categories, a topic we are going to discuss in more detail in Section \ref{sec:disco}. \cite{Baroni} composition method for adjectives and nouns also follows the very same principle.

\subsection{Deep learning models}
\label{sec:deeplearning}

\begin{figure}[h]
  \centering
  \includegraphics[scale=0.9]{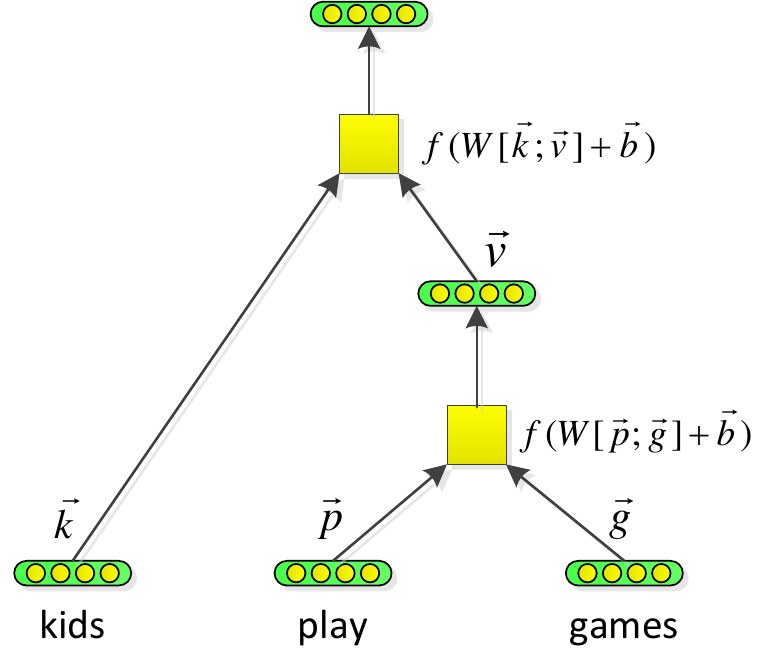}
  \caption{A recursive neural network with a single layer for providing compositionality in distributional models.}
  \label{fig:nn}
\end{figure}

A recent trend in compositionality of distributional models is based on \textit{deep learning} techniques, a class of machine learning algorithms (usually neural networks) that approach models as multiple layers of representations, where the higher-level concepts are induced from the lower-level ones. For example, \cite{socher2010,socher2011,socher2012} use recursive neural networks in order to produce compositional vector representations for sentences, with very promising results in a number of tasks. In its most general form, a neural network like this takes as input a pair of word vectors $\ov{w_1},\ov{w_2}$ and returns a new composite vector $\ov{y}$ according to the following equation:

\begin{equation}
  \ov{y} = g(\mathbf{W}[\ov{w}_1;\ov{w}_2] + \ov{b})
\end{equation}

\noindent where $[\ov{w}_1;\ov{w}_2]$ denotes the concatenation of the two child vectors, $\mathbf{W}$ and $\ov{b}$ are the parameters of the model, and $g$ is a non-linear function such as $\tanh$. This output, then, will be used in a recursive fashion again as input to the network for computing the vector representations of larger constituents. The general architecture of this model is presented in Figure \ref{fig:nn}.

Neural networks can vary in design and topology. \cite{nal13cvs}, for example, model sentential compositionality using a \textit{convolutional} neural network in an element-wise fashion. Specifically, the input of the network is a vector representing a single feature, the elements of which are collected across all the word vectors in the sentence. Each layer of the network applies convolutions of kernels of increasing size, producing at the output a single value that will form the corresponding feature in the resulting sentence vector. This method was used for providing sentence vectors in the context of a discourse model, and was tested with success in a task of recognizing dialogue acts of utterance within a conversation. Furthermore, it has been used as a sentence generation apparatus in a machine translation model with promising results \citep{nal13emnlp}.

On the contrary to the previously discussed compositional approaches, deep learning methods are based on a large amount of pre-training: the parameters $\mathbf{W}$ and $\ov{b}$ in the network of Figure \ref{fig:nn} must be learned through an iterative algorithm known as \textit{backpropagation}, a process that can be very time-consuming and in general cannot guarantee optimality. However, the non-linearity in combination with the layered approach in which neural networks are based provide these models with great power, allowing them to simulate the behaviour of a range of functions much wider than the linear maps of tensor-based approaches. Indeed, the work of  \cite{socher2010,socher2011,socher2012} has been tested in various paraphrase detection and sentiment analysis tasks, delivering results that by the time of this writing remain state-of-the-art.

\section{A categorical framework for natural language}
\label{sec:disco}

Tensor-based models stand in between the two extremes of vector mixtures and deep learning methods, offering an appealing alternative that can be powerful enough and at the same time fully aligned with the formal semantics view of natural language. Actually, it has been shown that the linear-algebraic formulas for the composite meanings produced by a tensor-based model emerge as the natural consequence of a structural similarity between a grammar and finite-dimensional vector spaces. In this section I will review the most important points of this work.

\subsection{Unifying grammar and meaning}
\label{sec:unifying}

Using the abstract mathematical framework of category theory, \cite{Coeckeetal} managed to equip the distributional models of meaning with compositionality in a way that every grammatical reduction corresponds to a linear map defining mathematical manipulations between vector spaces. In other words, given a sentence $s=w_1w_2\cdots w_n$ there exists a syntax-driven linear map $f$ from the context vectors of the individual words to a vector for the whole sentence defines as follows:

\begin{equation}
\label{equ:sent}
  \ov{s} = f(\ov{w_1} \otimes \ov{w_2} \otimes \cdots \otimes \ov{w_n})
\end{equation}

This result is based on the fact that the base type-logic of the framework, a pregroup grammar \citep{Lambek}, shares the same abstract structure with finite-dimensional vector spaces, that of a compact closed category. Mathematically, the transition from grammar types to vector spaces has the form of a strongly monoidal functor, that is, of a map that preserves the basic structure of compact closed categories. The following section provides a short introduction to the category theoretic notions above.

\subsection{Introduction to categorical concepts}

\textit{Category theory} is an abstract area of mathematics, the aim of which is to study and identify universal properties of mathematical concepts that often remain hidden by traditional approaches. A \textit{category} is a collection of objects and morphisms that hold between these objects, with composition of morphisms as the main operation. That is, for two morphisms $A\xrightarrow{f}B\xrightarrow{g}C$, we have $g \circ f: A\xrightarrow{}C$. Morphism composition is associative, so that $(f \circ g) \circ h = f \circ (g \circ h)$. Furthermore, every object $A$ has an identity morphism $1_A:A \to A$; for $f:A \to B$ we moreover require that:

\begin{equation}
 f \circ 1_A = f~~~\text{and}~~~ 1_B \circ f = f
\end{equation}

A \textit{monoidal category} is a special type of category equipped with another associative operation, a monoidal tensor $\otimes: \mathcal{C} \times \mathcal{C} \to \mathcal{C}$ (where $\mathcal{C}$ is our basic category). Specifically, for each pair of objects $(A,B)$ there exists a composite object $A\otimes B$, and for every pair of morphisms $(f:A \to C$, $g:B \to D)$ a parallel composite $f\otimes g: A \otimes B \to C \otimes D$. For a \textit{symmetric monoidal category}, it is also the case that $A \otimes B \cong B \otimes A$. Furthermore, there is a unit object $I$ which satisfies the following isomorphisms:

\begin{equation}
\label{equ:moniso}
 A\otimes I \cong A \cong I\otimes A
\end{equation}

A monoidal category is \textit{compact closed} if every object $A$ has a left and right adjoint, denoted as $A^l,A^r$ respectively, for which the following special morphisms exist:

\begin{equation}
\label{equ:eta}
  \eta^l: I \to A \otimes A^l~~~~
  \eta^r:I \to A^r \otimes A
\end{equation}
\vspace{-0.5cm}
\begin{equation}
\label{equ:epsilon}
  \epsilon^l: A^l \otimes A \to I~~~~
  \epsilon^r: A \otimes A^r \to I
\end{equation}

For the case of a \textit{symmetric compact closed category}, the left and right adjoints collapse into one, so that $A^* := A^l = A^r$.

Both a pregroup grammar and the category of finite-dimensional vector spaces over a base field $\mathbb{R}$ conform to this abstract definition of a compact closed category. A \textit{pregroup grammar} \citep{Lambek} is a type-logical grammar built on the rigorous mathematical basis of a pregroup algebra, i.e. a partially ordered monoid with unit $1$, whose each element $p$ has a left adjoint $p^l$ and a right adjoint $p^r$. These elements of the monoid are the objects of the category, the partial orders are the morphisms, the pregroup adjoints correspond to the adjoints of the category, while the monoid multiplication is the tensor product with 1 as unit. In a context of a pregroup, Equations \ref{equ:eta} and \ref{equ:epsilon} are transformed to the following:

\begin{equation}
\label{equ:pregroups}
  p^l  p \leq 1 \leq p  p^l~~~~\text{and}~~~~p  p^r \leq 1 \leq p^r  p
\end{equation}

\noindent where the juxtaposition of elements denotes the monoid multiplication. Each element $p$ represents an atomic type of the grammar, for example $n$ for noun phrases and $s$ for sentences. Atomic types and their adjoints can be combined to form compound types, e.g. $n^rsn^l$ for a transitive verb. This type reflects the fact that a transitive verb is an entity expecting for a noun at its right (the object) and a noun at its left (the subject) in order to return a sentence. The rules of the grammar are prescribed by the mathematical properties of pregroups, and specifically by the inequalities in (\ref{equ:pregroups}) above. A partial order in the context of a logic denotes implication, so from (\ref{equ:pregroups}) we derive:

\begin{equation}
\label{equ:pregr}
  p^lp \to 1~~~~\text{and}~~~~pp^r \to 1
\end{equation}

These cancellation rules to the unit object correspond to the $\epsilon$-maps of a compact closed category. It also holds that $1p = p = p1$, satisfying the isomorphism of the monoidal unit in (\ref{equ:moniso}). A derivation in a pregroup grammar has the form of a reduction diagram, where the cancellation rules (i.e. $\epsilon$-maps) are depicted by lines connecting elements:

\begin{exe}
\ex 
\begin{tikzpicture}
	\begin{pgfonlayer}{nodelayer}
		\node [style=none, text height=1.5 ex, text depth=0.25 ex] (0) at (-7.5, 1.75) {John};
		\node [style=none, text height=1.5 ex, text depth=0.25 ex] (1) at (-5, 1.75) {saw};
		\node [style=none, text height=1.5 ex, text depth=0.25 ex] (2) at (-2.5, 1.75) {Mary};
		\node [style=none, text height=1.5 ex, text depth=0.25 ex] (3) at (0.25, 1.75) {reading};
		\node [style=none, text height=1.5 ex, text depth=0.25 ex] (4) at (3.1, 1.75) {a};
		\node [style=none, text height=1.5 ex, text depth=0.25 ex] (5) at (5, 1.75) {book};
		\node [style=none, text height=1.5 ex, text depth=0.25 ex] (6) at (-7.5, 0.75) {$n$};
		\node [style=none, text height=1.5 ex, text depth=0.25 ex] (7) at (-5.75, 0.75) {$n^r$};
		\node [style=none, text height=1.5 ex, text depth=0.25 ex] (8) at (-5, 0.75) {$s$};
		\node [style=none, text height=1.5 ex, text depth=0.25 ex] (9) at (-4.25, 0.75) {$n^l$};
		\node [style=none, text height=1.5 ex, text depth=0.25 ex] (10) at (-2.5, 0.75) {$n$};
		\node [style=none, text height=1.5 ex, text depth=0.25 ex] (11) at (-0.5, 0.75) {$n^r$};
		\node [style=none, text height=1.5 ex, text depth=0.25 ex] (12) at (0.25, 0.75) {$n$};
		\node [style=none, text height=1.5 ex, text depth=0.25 ex] (13) at (1, 0.75) {$n^l$};
		\node [style=none, text height=1.5 ex, text depth=0.25 ex] (14) at (2.75, 0.75) {$n$};
		\node [style=none, text height=1.5 ex, text depth=0.25 ex] (15) at (3.4, 0.75) {$n^l$};
		\node [style=none, text height=1.5 ex, text depth=0.25 ex] (16) at (5, 0.75) {$n$};
		\node [style=none] (17) at (-7.5, 0.25) {};
		\node [style=none] (18) at (-5.75, 0.25) {};
		\node [style=none] (19) at (-5, 0.25) {};
		\node [style=none] (20) at (-4.25, 0.25) {};
		\node [style=none] (21) at (-2.5, 0.25) {};
		\node [style=none] (22) at (-0.5, 0.25) {};
		\node [style=none] (23) at (0.25, 0.25) {};
		\node [style=none] (24) at (1, 0.25) {};
		\node [style=none] (25) at (2.75, 0.25) {};
		\node [style=none] (26) at (3.5, 0.25) {};
		\node [style=none] (27) at (5, 0.25) {};
		\node [style=none] (28) at (-5, -0.75) {};
	\end{pgfonlayer}
	\begin{pgfonlayer}{edgelayer}
		\draw [thick, bend left=270] (21.center) to (22.center);
		\draw [thick, bend left=270] (24.center) to (25.center);
		\draw [thick, bend left=270, looseness=1.25] (26.center) to (27.center);
		\draw [thick, looseness=0.00] (19.center) to (28.center);
		\draw [thick, bend right=90, looseness=0.75] (20.center) to (23.center);
		\draw [thick, bend right=90] (17.center) to (18.center);
	\end{pgfonlayer}
\end{tikzpicture}}

\end{exe}

We refer to the compact closed category formed by a pregroup freely generated over a set of basic types (here $\{n,s\}$) by \textbf{Preg_F}. On the other hand, finite-dimensional vector spaces also form a compact closed category, where the vector spaces are the objects, linear maps are the morphisms, the main operation is the tensor product between vector spaces, and the field over which the spaces are formed, in our case $\mathbb{R}$, is the unit. The adjoints are the dual spaces, which are isomorphic to the space itself (hence, on the contrary with \textbf{Preg_F}, finite-dimensional vector spaces form a \textit{symmetric} compact closed category); the $\epsilon$-maps correspond to the inner product between the involved context vectors, as follows:

\begin{equation}
\label{equ:innerprod}
  \epsilon^l = \epsilon^r: W \otimes W \to \mathbb{R}:: \sum\limits_{ij} c_{ij}(\ov{w_i} \otimes \ov{w_j}) \mapsto \sum\limits_{ij} c_{ij}\langle \ov{w_i}|\ov{w_j}\rangle
\end{equation}

Let us refer to the category of finite-dimensional vector spaces and linear maps over a field as \textbf{FVect_W}, where $W$ is our basic distributional vector space with an orthonormal basis $\{w_i\}_i$. The transition from grammar type reductions to vector spaces, then, is accomplished by a \textit{strongly monoidal functor} $\mathcal{F}$ of the form:

\begin{equation}
 \mathcal{F}: \textbf{Preg_F} \to \textbf{FVect_W}
\end{equation}

\noindent
which preserves the compact structure between the two categories so that $\mathcal{F}(A^l) = \mathcal{F}(A)^l$ and $\mathcal{F}(A^r) = F(A)^r$. The categorical framework is agnostic regarding the form of our sentence space $S$, so in general $\mathcal{F}(n)$ can be different than $\mathcal{F}(s)$. However, in order to keep this presentation simple let us have our functor assigning the same basic vector space $W$ to both of the basic types, as follows:

\begin{equation}
  \label{equ:map}
  \mathcal{F}(n) = \mathcal{F}(s) = W 
\end{equation}

\noindent
Furthermore, the complex types are mapped to tensor products of vector spaces:

\begin{equation}
 \label{equ:complmap}
 \mathcal{F}(nn^l) = \mathcal{F}(n^rn) = W \otimes W~~~~~~~
 \mathcal{F}(n^rsn^l) = W \otimes W \otimes W
\end{equation}

Similarly, the type reductions are mapped to the compositions of tensor products of identity and $\epsilon$-maps of \textbf{FVect_W}. I will use the case of a transitive sentence as an example. Here, the subject and the object have the type $n$, whereas the type of the verb is $n^rsn^l$, as described above. The derivation proceeds as follows:

\[
  n(n^rsn^l)n = (nn^r)s(n^ln) \to 1s1 = s
\]

\noindent
which corresponds to the following map: 

\begin{align}
  \mathcal{F}(n(n^rsn^l)n) &= \mathcal{F}(\epsilon_n^r \otimes 1_s \otimes \epsilon_n^l) \\ \nonumber
  &= \epsilon_W \otimes 1_W \otimes \epsilon_W: W \otimes (W\otimes W \otimes W) \otimes W \to W
\end{align}

The function $f$ in Equation \ref{equ:sent}, suggested as a way for calculating the meaning of a sentence, now takes a concrete form strictly based on the grammar rules that connect the individual words. Let us work on the transitive example a bit further: the map of pregroup types to tensors prescribes that the subject and object are vectors (Eq. \ref{equ:map}) while the verb is a tensor of order 3 (Eq. \ref{equ:complmap}). So for a simple sentence such as `dogs chase cats' we have the following geometric entities:

\[
\ov{dogs} = \sum_i c^{dogs}_i \ov{w_i}~~~~~~~~~\overline{chase} = \sum_{ijk}c^{chase}_{ijk}(\ov{w_i} \otimes \ov{w_j} \otimes \ov{w_k})~~~~~~~~~\ov{cats} = \sum_k c^{cats}_k \ov{w_k}
\]

\noindent where $c^v_i$ denotes the $i$th component in vector $\ov{v}$ and $\ov{w_i}$ a basis vector of $W$. Applying Equation \ref{equ:sent} on this sentence will give:

\begin{equation}
\label{deriv}
   \begin{split}
   & \mathcal{F}(\epsilon^r_n \otimes 1_n \otimes \epsilon^l_n)(\ov{dogs} \otimes \overline{chase} \otimes \ov{cats}) = \\
   & ~~~~~~ = (\epsilon_W \otimes 1_W \otimes \epsilon_W)(\ov{dogs} \otimes \overline{chase} \otimes \ov{cats}) = \\
   & ~~~~~~ = \sum\limits_{ijk} c_{ijk}^{chase} \langle \ov{dogs}|\ov{w_i} \rangle \ov{w_j} \langle \ov{w_k}|\ov{cats}\rangle = \\
   & ~~~~~~ = \ov{dogs}^{\mathsf{T}} \times \overline{chase} \times \ov{cats}
   \end{split}
\end{equation}

\noindent where the symbol $\times$ denotes tensor contraction. Thus we have arrived at Equation \ref{equ:trans}, presented in Section \ref{sec:tensorbased} as a means for calculating the vector of a transitive sentence in the context of a tensor-based model. 

The significance of the categorical framework lies exactly in this fact, that it provides an elegant mathematical counterpart of the formal semantics perspective as expressed by \cite{Mon1}, where words are represented and interact with each other according to their type-logical identities. Furthermore, it seems to imply that approaching the problem of compositionality in a tensor-based setting is a step towards the right direction, since the linear-algebraic manipulations come as a direct consequence of the grammatical derivation. 
The framework itself is a high-level recipe for composition in distributional environments, leaving a lot of room for further research and experimental work. Concrete implementations have been provided, for example, by \cite{GrefenSadr1} and \cite{kartsaklis2012}. For more details on pregroup grammars and their type dictionary, see \cite{Lambek}. The functorial passage from a pregroup grammar to finite-dimensional vector spaces is described in detail in \cite{kartsaklis2013}.

\section{Verb and sentence spaces}
\label{sec:sentencespace}

Until now I have been deliberately vague when talking about the sentence space $S$ and the properties that a structure like this should bring. In this section I will try to discuss this important issue in more detail, putting some emphasis on how it is connected to another blurry aspect of the discussion so far, the form of relational words such as verbs. From a mathematical perspective, the decisions that need to be taken regard (a) the dimension of $S$, that is, the cardinality of its basis; and (b) the form of the basis. In other words, how many and what kind of features will comprise the meaning of a sentence? 

This question finds a trivial answer in the setting of vector mixture models; since everything in that approach lives into the same base space, a sentence vector has to share the same size and features with words. It is instructive to pause for a moment and consider what does this really mean in practice. What a distributional vector for a word actually shows us is to what extent all other words in the vocabulary are related to this specific target word. If our target word is a verb, then the components of its vector can be thought as related to the \textit{action} described by the verb: a vector for  the verb `run' reflects the degree to which a `dog' can run, a `car' can run, a `table' can run and so on. The element-wise mixing of vectors $\ov{dog}$ and $\ov{run}$ then in order to produce a compositional representation for the meaning of the simple intransitive sentence `dogs run', finds an intuitive interpretation: the output vector will reflect the extent to which things that are related to dogs can also run; in other words, it shows how \textit{compatible} the verb is with the specific subject. 

A tensor-based model, on the other hand, goes beyond a simple compatibility check between the relational word and its arguments; its purpose is to \textit{transform} the noun into a sentence. Furthermore, the size and the form of the sentence space become tunable parameters of the model, which can depend on the specific task in hand. Let us assume that in our model we select sentence and noun spaces such that $S \in \mathbb{R}^s$ and $N \in \mathbb{R}^n$, respectively; here, $s$ refers to the number of distinct features that we consider appropriate for representing the meaning of a sentence in our model, while $n$ is the corresponding number for nouns. An intransitive verb, then, like `play' in `kids play', will live in $N \otimes S \in \mathbb{R}^{n \times s}$, and will be a map $f: N \to S$ built (somehow) in a way to take as input a noun and produce a sentence; similarly, a transitive verb will live in $N \otimes S \otimes N \in \mathbb{R}^{n \times s \times n}$ and will correspond to a map $f_{tr}: N\otimes N \to S$. We can now provide some intuition of how the verb space should be linked to the sentence space: the above description clearly suggests that, in a certain sense, the verb tensor should be able to somehow encode the meaning of \textit{every} possible sentence that can be produced by the specific verb, and emit the one that matches better the given input. 


Let us demonstrate these ideas using a concrete example, where the goal is to simulate the truth-theoretic nature of formal semantics view in the context of a tensor-based model (perhaps for the purposes of a textual entailment task). In that case, our sentence space will be nothing more than the following:

\singlespace
\begin{equation}
S = 
\left\lbrace
\left( \begin{array}{c}
0 \\
1  
\end{array} \right)
,
\left( \begin{array}{c}
1 \\
0  
\end{array} \right)
\right\rbrace
\end{equation}

\noindent
with the two vectors representing $\top$ and $\bot$, respectively. Each individual in our universe will correspond to a basis vector of our noun space; with just three individuals (Bob, John, Mary), we get the following mapping:

\singlespace
\begin{equation}
\ov{bob} = 
\left( \begin{array}{c}
0 \\
0 \\
1  
\end{array} \right)
~~~\ov{john}=
\left( \begin{array}{c}
0 \\
1 \\
0  
\end{array} \right)
~~~\ov{mary}=
\left( \begin{array}{c}
1 \\
0 \\
0  
\end{array} \right)
\end{equation}

In this setting, an intransitive verb will be a matrix formed as a series of truth values, each one of which is associated with some individual in our universe. Assuming for example that only John performs the action of sleeping, then the meaning of the sentence `John sleeps' is given by the following computation:

\singlespace
\begin{equation}
\ov{john}^{\mathsf{T}} \times \overline{sleep} =
\left( \begin{array}{cccc}
0 & 1 & 0 
\end{array} \right)
\times
\left( \begin{array}{cccc}
1 & 0 \\
0 & 1 \\
1 & 0 \\
\end{array} \right)
=
\left( \begin{array}{c}
0 \\
1  
\end{array} \right)
=
\top
\end{equation}

The matrix for \textit{sleep} provides a clear intuition of a verb structure that encodes all possible sentence meanings that can be emitted from the specific verb. Notice that each row of $\overline{sleep}$ corresponds to a potential sentence meaning given a different subject vector; the role of the subject is to specify which row should be selected and produced as the output of the verb. 

It is not difficult to imagine how this situation scales up when we move on from this simple example to the case of highly-dimensional real-valued vector spaces. Regarding the first question we posed in the beginning of this section, this implies some serious practical limitations on how large $s$ can be. Although it is generally true that the higher the dimension of sentence space, the subtler the differences we would be able to detect from sentence to sentence, in practice a verb tensor must be able to fit in a computer's memory to be of any use to us; with today's machines, this could roughly mean that $s \leq 300$.\footnote{To understand why, imagine that a ditransitive verb is a tensor of order 4 (a ternary function); by taking $s=n=300$ this means that the required space for just one ditransitive verb would be $300^4 \times$ 8 bytes per number $\approx$ 65 gigabytes.} Using very small vectors is also common practice in deep learning approaches, aiming to reduce the training times of the expensive optimization process; \cite{socher2012}, for example, use 50-dimensional vectors for both words and phrases/sentences.

The second question posed in this section, regarding \textit{what kind} of properties should comprise the meaning of a sentence, seems more philosophical than technical. Although in principle model designers are free to select whatever features (that is, basis vectors of $S$) they think might serve better the purpose of the model, in practice an empirical approach is usually taken in order to sidestep deeper philosophical issues regarding the nature of a sentence. In a deep learning setting, for example, the sentence vector emerges as a result of an \textit{objective function} based on which the parameters of the model have been optimized. In \citep{socher2011}, the objective function assesses the quality of a parent vector (e.g. vector $\ov{v}$ in Figure \ref{fig:nn}) by how faithfully it can be deconstructed to the two original children vectors. Hence, the sentence vector at the top of the diagram in Figure \ref{fig:nn} is a vector constructed in a way to allow the optimal reconstruction of the vectors for its two children, the noun phrase `kids' and the verb phrase `play games'. The important point here is that no attempt has been made to interpret the components of the sentence vector individually; the only thing that matters is how faithfully the resulting vector fulfils the adopted constraints.

In the context of a tensor-based model, the question regarding the form of a sentence space can be recast in the following form: How should we build the sentence-producing maps (i.e. our verbs) in order to output the appropriate form of sentence vector or tensor? \cite{Baroni} propose a method for building adjective matrices, which is in fact directly applicable to any relational word. Assuming we want to create a matrix for the intransitive verb `run', we can collect all instances of this verb occurring together with some subject in the corpus and create a distributional vector for these two-word constructs based on their contexts as if they were single words; each one of these vectors is paired with the vector of the corresponding subject to create a set of the form: $\langle \ov{\textit{dog}}$, $\ov{\textit{dogs run}} \rangle, \langle \ov{\textit{people}}$, $\ov{\textit{people run}} \rangle, \langle \ov{\textit{car}}, \ov{\textit{cars run}} \rangle$ and so on. We can now use linear regression in order to produce an appropriate matrix for `run' based on these exemplars. Specifically, the goal of the learning process is to find the matrix $\overline{run}$ that minimizes the following quantity:

\begin{equation}
  \sum\limits_i \left( \overline{run} \times subj_i - \ov{subj_i~run} \right)^2
\end{equation}

\noindent which represents the total error for all nouns occurred as subjects of the specific intransitive verb. Notice that this time our objective function has a different form from the reconstruction error of Socher et al., but it still exists: the verb must be able to produce a sentence vector that, given an arbitrary subject, will approximate the distributional behaviour of all those two-word elementary exemplars on which the training was based.

\section{Challenges and open questions}
\label{sec:challenges}

The previous sections hopefully provided a concise introduction to the important developments that have been noted in the recent years on the topic of compositional-distributional models. Despite this progress, however, the provision of distributional models with compositionality is an endeavour that still has a long way to go. This section outlines some important issues that current and future research should face in order to provide a more complete account to the problem.

\subsection{Evaluating the correctness of distributional hypothesis}

The idea presented in Section \ref{sec:sentencespace} for creating distributional vectors of constructions larger than single words has its roots to an interesting thought experiment, the purpose of which is to investigate the potential distributional behaviour of large phrases and sentences and the extent to which such a distributional approach is plausible or not for longer-than-words text constituents. The argument goes like this: if we had an infinitely large corpus of text, we could create a purely distributional representation of the phrase or sentence, exactly as we do for words, by taking into account the different contexts within which this text fragment occurs. Then the assumption would be that the vector produced by the composition of the individual word vectors should be a synthetic ``reproduction'' of this distributional sentence vector. This thought experiment poses some interesting questions: First of all, it is not clear if the distributional hypothesis does indeed scale up to text constituents larger than words; second, even if we assume it does, what ``context'' would mean in this case? For example, what would an appropriate context be of a 20-word sentence?

Although there is no such a thing as an ``infinitely large corpus'', it would still be possible to get an insight about these important issues if we restrict ourselves to small constructs---say, two-word constituents---for which we can still get reliable frequency counts from a large corpus. In the context of their work with adjectives (shortly discussed in Section \ref{sec:sentencespace}), \cite{Baroni} performed an interesting experiment along these lines using the ukWaC corpus, consisted of 2.8 billion words. As we saw, their work follows the tensor-based paradigm where adjectives are represented as linear maps learnt using linear regression and act on the context vectors of nouns. The composite vectors were compared with observed vectors of adjective-noun compounds, created by the contexts of each compound in the corpus. The results, although perhaps encouraging, are far from perfect: for 25\% of the composed vectors, the observed vector was not even in the top 1,000 of their nearest neighbours, in 50\% of the cases the observed vector was in the top 170, while only for a 25\% of the cases the observed vector was in the top 17 of nearest neighbours.

As the authors point out, one way to explain the performance is as a result of data sparseness. However, the fact that a 2.8 billion-word corpus is not sufficient for modelling elementary two-word constructs would be really disappointing. As \cite{pulman2013} mentions:

\begin{quote}
``It is worth pointing out that a corpus of 2.83 billion is already thousands of times as big as the number of words it is estimated a 10-year-old person would have been exposed to \citep{moore2003}, and many hundreds of times larger than any person will hear or read in a lifetime.''
\end{quote}

If we set aside for a moment the possibility of data sparseness, then we might have to start worrying about the validity of our fundamental assumptions, i.e. that of distributional hypothesis and principle of compositionality. Does the result mean that the distributional hypothesis holds only for individual words such as nouns, but it is not very effective for larger constituents such as adjective-noun compounds? Doubtful, since a noun like `car' and an adjective-noun compound like `red car' represent similar entities, share the same structure, and occur within similar contexts. Is this then an indication that the distributional hypothesis suffers from some fundamental flaw that limits its applicability even for the case of single words? That would be a very strong and rather unjustified claim to make, since it is undoubtedly proven that distributional models can capture the meaning of words, at least to some extent, for many real-world applications (see examples in Section \ref{sec:word2sentence}). Perhaps we should seek the reason of the sub-optimal performance to the specific methods used for the composition in that particular experiment. The adjectives, for example, are modelled as linear functions over their arguments (nouns they modify), which raises another important question: Is \textit{linearity} an appropriate model for composition in natural language? Further research is needed in order to provide a clearer picture of the expectations we should have regarding the true potential of compositional-distributional models, and all the issues raised in this section are very important towards this purpose.

\subsection{What is ``meaning''?}

Even if we accept that the distributional hypothesis is correct, and a context vector can indeed capture the ``meaning'' of a word, it would be far too simplistic to assume that this holds for \textit{every} kind of word. The meaning of some words can be determined by their denotations; it is reasonable to claim, for example, that the meaning of the word `tree' is the set of all trees, and we are even able to answer the question ``what is a tree?'' by pointing to a member of this set, a technique known as \textit{ostensive} definition. But this is not true for all words. In ``Philosophy'' (published in ``Philosophical Occasions: 1912-1951'', \citeyear{witphilosophy}), Ludwig Wittgenstein notes that there exist certain words, like `time', the meaning of which is quite clear to us until the moment we have to explain it to someone else; then we realize that suddenly we are not able any more to express in words what we certainly know---it is like we have forgotten what that specific word really means. Wittgenstein claims that ``if we have this experience, then we have arrived at the limits of language''. This observation is related to one of the central ideas of his work: that the meaning of a word does not need to rely on some kind of definition; what really matters is the way we use this word in our everyday communications. In ``Philosophical Investigations'' (\citeyear{wittgenstein1963}), Wittgenstein presents a thought experiment:

\begin{quote}
``Now think of the following use of language: I send someone shopping. I give him a slip marked `five red apples'. He takes the slip to the shopkeeper, who opens the drawer marked `apples'; then he looks up the word `red' in a table and finds a colour sample opposite it; then he says the series of cardinal numbers---I assume that he knows
them by heart---up to the word `five' and for each number he takes an apple of the same colour as the sample out of the drawer. It is in this and similar ways that one operates with words.''
\end{quote}

For the shopkeeper, the meaning of words `red' and `apples' was given by ostensive definitions (provided by the colour table and the drawer label). But what was the meaning of word `five'? Wittgenstein is very direct on this:

\begin{quote}
``No such thing was in question here, only how the word `five' is used''.
\end{quote}

If language is not expressive enough to describe certain fundamental concepts of our world, then the application of distributional models is by definition limited. Indeed, the subject of this entire section is the concept of `meaning'---yet, how useful would be for us to use this resource in order to construct a context vector for this concept? To what extent would this vector be an appropriate semantic representation for the word `meaning'?

\subsection{Treatment of functional words}
\label{sec:functional}

In contrast with content words, like nouns, verbs, and adjectives, functional words such as prepositions, determiners, or relative pronouns are considered semantically vacuous. In the context of a distributional model, the problem arises from the ubiquitous nature of these words, which means that creating a context vector for preposition `in', for example, would not be especially useful, since this word can be encountered in almost every possible context. Vector mixture models ``address'' this problem by just ignoring all these functional words, a solution that seems questionable given that these words signify specific relations between different text constituents. Under the formal semantics view, for example, a noun phrase such as `man in uniform' would be represented by the following logical form:

\begin{exe}
  \ex $\exists x\exists y.[\textit{man}(x) \wedge \textit{uniform}(y) \wedge \textit{in}(x,y)]$
\end{exe}

\noindent
where the predicate $\textit{in}(x,y)$ denotes that between the two entities holds a specific kind of relation. At the other end of the scale, deep learning techniques rely on their brute force and make no distinction between functional words and content words, hoping that the learning process will eventually capture at least some of the correct semantics. To what extent this can be achieved, though, it is still not quite clear. 

Compositional models that are based on deep semantic analysis, as the categorical framework of \cite{Coeckeetal} is, do not allow us to arbitrarily drop a number of words from a sentence based on subjective criteria regarding their meaning, and for a good reason: there is simply no linguistic justification behind such decisions. However, since in that case a functional word is a multi-linear map, we can still manually ``engineer'' the inner workings of this structure in a way that conforms to the linguistic intuition we have for the role of the specific word within the sentence. For example, \cite{SadrRelpronouns} use Frobenius algebras over the category of finite-dimensional vector spaces in order to detail the ``anatomy'' of subjective and objective relative pronouns. Given that the type of a subjective relative pronoun is $n^rns^ln$, a typical derivation (here using a pregroup grammar) gets the following form:

\begin{exe}
\ex 
\begin{tikzpicture}[scale=0.80]
	\begin{pgfonlayer}{nodelayer}
		\node [style=none] (0) at (-6.75, 1.75) {dog};
		\node [style=none] (1) at (-2.25, 1.75) {that};
		\node [style=none] (2) at (2.75, 1.75) {bites};
		\node [style=none] (3) at (7, 1.75) {men};
		\node [style=none, text height=1.5 ex, text depth=0.25 ex] (4) at (-6.75, 0.5) {$n$};
		\node [style=none, text height=1.5 ex, text depth=0.25 ex] (5) at (-3.5, 0.5) {$n^r$};
		\node [style=none, text height=1.5 ex, text depth=0.25 ex] (6) at (-2.5, 0.5) {$n$};
		\node [style=none, text height=1.5 ex, text depth=0.25 ex] (7) at (-1.75, 0.5) {$s^l$};
		\node [style=none, text height=1.5 ex, text depth=0.25 ex] (8) at (-0.75, 0.5) {$n$};
		\node [style=none, text height=1.5 ex, text depth=0.25 ex] (9) at (1.75, 0.5) {$n^r$};
		\node [style=none, text height=1.5 ex, text depth=0.25 ex] (10) at (2.75, 0.5) {$s$};
		\node [style=none] (11) at (2.75, 0.5) {};
		\node [style=none, text height=1.5 ex, text depth=0.25 ex] (12) at (3.75, 0.5) {$n^l$};
		\node [style=none, text height=1.5 ex, text depth=0.25 ex] (13) at (7, 0.5) {$n$};
		\node [style=none] (14) at (-6.75, -0.25) {};
		\node [style=none] (15) at (-3.5, -0.25) {};
		\node [style=none] (16) at (-2.5, -0.25) {};
		\node [style=none] (17) at (-1.75, -0.25) {};
		\node [style=none] (18) at (-0.75, -0.25) {};
		\node [style=none] (19) at (1.75, -0.25) {};
		\node [style=none] (20) at (2.75, -0.25) {};
		\node [style=none] (21) at (3.75, -0.25) {};
		\node [style=none] (22) at (7, -0.25) {};
		\node [style=none] (23) at (-2.5, -2.25) {};
	\end{pgfonlayer}
	\begin{pgfonlayer}{edgelayer}
		\draw [thick, bend right=90, looseness=1.25] (14.center) to (15.center);
		\draw [thick] (16.center) to (23.center);
		\draw [thick, bend right=90, looseness=1.25] (17.center) to (20.center);
		\draw [thick, bend left=270, looseness=1.50] (18.center) to (19.center);
		\draw [thick, bend left=270, looseness=1.25] (21.center) to (22.center);
	\end{pgfonlayer}
\end{tikzpicture}}

\end{exe}

In these cases, the part that follows the relative pronoun acts as a modifier on the head noun. It is possible, then, to define the inner structure of tensor $\overline{that}$ in a way that allows us to pass the information of the noun from the left-hand part of the phrase to the right-hand part, in order to let it properly interact with the modifier part. Schematically, this is depicted by:

\begin{exe}
\ex 
\begin{tikzpicture}[scale=0.80]
	\begin{pgfonlayer}{nodelayer}
		\node [style=none] (0) at (-3.5, 2) {};
		\node [style=none] (1) at (-2.75, 2) {};
		\node [style=none] (2) at (-1.75, 2) {};
		\node [style=none] (3) at (-0.25, 2) {};
		\node [style=none] (4) at (-6.75, 1.75) {dog};
		\node [style=none, circle, minimum size=0.15 cm, fill=white, draw] (5) at (-2.25, 1.75) {};
		\node [style=none, circle, minimum size=0.15 cm, fill=white, draw] (6) at (-1.25, 1.75) {};
		\node [style=none] (7) at (3.25, 1.75) {bites};
		\node [style=none] (8) at (7.5, 1.75) {men};
		\node [style=none] (9) at (-3.5, 1) {};
		\node [style=none] (10) at (-2.25, 1) {};
		\node [style=none] (11) at (-1.25, 1) {};
		\node [style=none] (12) at (-0.25, 1) {};
		\node [style=none, text height=1.5 ex, text depth=0.25 ex] (13) at (-6.75, 0.5) {$n$};
		\node [style=none, text height=1.5 ex, text depth=0.25 ex] (14) at (-3.5, 0.5) {$n^r$};
		\node [style=none, text height=1.5 ex, text depth=0.25 ex] (15) at (-2.25, 0.5) {$n$};
		\node [style=none, text height=1.5 ex, text depth=0.25 ex] (16) at (-1.25, 0.5) {$s^l$};
		\node [style=none, text height=1.5 ex, text depth=0.25 ex] (17) at (-0.25, 0.5) {$n$};
		\node [style=none, text height=1.5 ex, text depth=0.25 ex] (18) at (2.25, 0.5) {$n^r$};
		\node [style=none, text height=1.5 ex, text depth=0.25 ex] (19) at (3.25, 0.5) {$s$};
		\node [style=none] (20) at (3.25, 0.5) {};
		\node [style=none, text height=1.5 ex, text depth=0.25 ex] (21) at (4.25, 0.5) {$n^l$};
		\node [style=none, text height=1.5 ex, text depth=0.25 ex] (22) at (7.5, 0.5) {$n$};
		\node [style=none] (23) at (-6.75, -0.25) {};
		\node [style=none] (24) at (-3.5, -0.25) {};
		\node [style=none] (25) at (-2.25, -0.25) {};
		\node [style=none] (26) at (-1.25, -0.25) {};
		\node [style=none] (27) at (-0.25, -0.25) {};
		\node [style=none] (28) at (2.25, -0.25) {};
		\node [style=none] (29) at (3.25, -0.25) {};
		\node [style=none] (30) at (4.25, -0.25) {};
		\node [style=none] (31) at (7.5, -0.25) {};
		\node [style=none] (32) at (-2.25, -2.25) {};
	\end{pgfonlayer}
	\begin{pgfonlayer}{edgelayer}
		\draw [thick] (11.center) to (6.center);
		\draw [thick, bend left=90, looseness=2.00] (0.center) to (1.center);
		\draw [thick] (3.center) to (12.center);
		\draw [thick, bend left=255] (1.center) to (2.center);
		\draw [thick, bend left=90] (2.center) to (3.center);
		\draw [thick, bend right=90, looseness=1.25] (23.center) to (24.center);
		\draw [thick] (25.center) to (32.center);
		\draw [thick] (5.center) to (10.center);
		\draw [thick, bend right=90, looseness=1.25] (26.center) to (29.center);
		\draw [thick, bend left=270, looseness=1.50] (27.center) to (28.center);
		\draw [thick, bend left=270, looseness=1.25] (30.center) to (31.center);
		\draw [thick] (0.center) to (9.center);
	\end{pgfonlayer}
\end{tikzpicture}}

\end{exe}

In the above diagram, the ``cups'' ($\cup$) correspond to $\epsilon$-maps as usual, the ``caps'' ($\cap$) represent $\eta$-maps (see Section \ref{sec:disco} and Equation \ref{equ:eta}), and the dots denote Frobenius operations. Explaining the exact manipulations that take place here is beyond the scope of this article, but intuitively the following things happen: 

\begin{enumerate}
  \item The sentence dimension of the verb is ``killed''; this can be seen as a collapse of the order-3 tensor into a tensor of order 2 along the sentence dimension.
  \item The new version of verb, now retaining information only from the dimensions linked to the subject and the object, interacts with the object and produces a new vector.
  \item This new vector is ``merged'' with the subject, in order to modify it appropriately.
\end{enumerate}

From a linear-algebraic perspective, the meaning of the noun phrase is given as the computation $(\overline{verb} \times \ov{obj}) \odot \ov{subj}$, where $\times$ denotes matrix multiplication and $\odot$ point-wise multiplication. Note that this final equation does not include any tensor for `that'; the word solely acts as a ``router'', moving information around and controlling the interactions between the content words. The above treatment of relative pronouns is flexible and possibly opens a door for modelling other functional words, such as prepositions. What remains to be seen is how effective can be in an appropriate large-scale evaluation, since the current results, although promising, are limited to a small example dataset.

\subsection{Treatment of logical words}
\label{sec:logical}

Logical words, like `not', `and ', `or', constitute a different category of functional words that need to be addressed. In the context of distributional semantics, a natural tool for achieving this is \textit{quantum logic}, originated by \cite{birkhoff}, where the logical connectives operate on linear subspaces of a vector space. Under this setting, the negation of a subspace is given by its orthogonal complement. Given two vector spaces $A$ and $V$, with $A$ to be a subspace of $V$ (denoted by $A \leq V$), the orthogonal complement of $A$ is defined as follows:

\begin{equation}
  A^\perp = \{ \ov{v} \in V: \forall \ov{a} \in A, \langle \ov{a} | \ov{v} \rangle = 0 \}
\end{equation}

The negation of a word vector $\ov{w}$, then, can be defined as the orthogonal complement of the vector space $\langle w \rangle=\{\lambda \ov{w}: \lambda \in \mathbb{R} \}$ generated by $\ov{w}$. \cite{widdows2003orthogonal} applies this idea on keywords, in the context of an information retrieval query. Specifically, he models negation by using projection to the orthogonal subspace, with very good results compared to traditional boolean operators. Suppose for example that we need to include in our query an ambiguous word like `rock', but we also wish to restrict its meaning to the geological sense (so we want the vector space counterpart of the boolean query \texttt{rock NOT music}). We can achieve this by projecting the vector for `rock' onto the orthogonal subspace of the vector for `music', an operation that eventually will retain the components of the vector of `rock' that are not related to music. For two arbitrary words $w_1$ and $w_2$, this can be achieved as follows:

\begin{equation}
  \ov{w_1} \wedge \neg \ov{w_2} = \frac{\langle \ov{w_1} | \ov{w_2} \rangle}{|\ov{w_2}|^2} \ov{w_2}
\end{equation}

In quantum logic, disjunction is modelled as the vector sum of subspaces: $V+W$ is the smallest subspace that includes both $V$ and $W$. That is, an expression like $w_1~OR~w_2$ is represented by the subspace:

\begin{equation}
  \langle w_1 \rangle + \langle w_2 \rangle = \{\lambda_1 \ov{w_1} + \lambda_2 \ov{w_2}:\lambda_1,\lambda_2 \in \mathbb{R} \}
\end{equation}

\noindent 
where again $\langle w \rangle$ is the vector space generated by vector $\ov{w}$. Finally, the conjunction of two subspaces is their intersection, that is, the set of vectors belonging to both of them. \cite{widdows2003orthogonal} discusses some of the subtleties of a practical implementation based on these concepts, such as difficulties on evaluating the similarity of a vector with a subspace. Furthermore, it should be mentioned that quantum logic has some important differences from classical logic, notably it does not adhere to the distributive law, so in general:

\begin{equation}
 A \wedge (B \vee C) \neq (A \wedge B) \vee (A \wedge C)
\end{equation}

The failure of distributivity law produces another unwelcome side-effect: quantum logic has no way to represent material implication, which in the context of a propositional logic is given by the following rule:

\begin{equation}
  P \to Q \Leftrightarrow \neg P \vee Q
\end{equation}

An excellent introduction to quantum logic, specifically oriented to information retrieval, can be found in \cite{rijsbergen}. 

\subsection{Quantification}

Perhaps the single most important obstacle for constructing proper compositional vectors for phrases and sentences is quantification---a concept that is incompatible with distributional models. A quantifier operates on a number of individual entities by counting or enumerating them: \textit{all} men, \textit{some} women, \textit{three} cats, \textit{at least} four days. This fits nicely in the logical view of formal semantics. Consider for example the sentence `John loves every woman', which has the following logical form:

\begin{exe}
  \ex $\forall x (\textit{woman}(x) \to \textit{loves}(john,x))$
\end{exe}

Given that the set of women in our universe is $\{\textit{alice},\textit{mary},\textit{helen}\}$, the above expression will be true if relation $loves$ includes the pairs $(john,alice),(john,mary),(john,helen)$. Unfortunately, this treatment does not make sense in vector space models since they lack any notion of individuality; furthermore, creating a context vector for a quantifier is meaningless. As a result, quantifiers are just considered ``noise'' and ignored in the current practice, producing unjustified equalities such as $\ov{every~man}=\ov{some~man}=\ov{man}$. The extent to which vector spaces can be equipped with quantification (and whether this is possible or not) remains another open question for further research. \cite{preller2013a} provides  valuable insights on the topic of moving from functional to distributional models and how these two approaches are related.

\section{Some closing remarks}

Compositional-distributional models of meaning constitute a technology with great potential that can drastically influence and improve the practice of natural language processing. Admittedly our efforts are still in their infancy, which should be evident from the discussion in Section \ref{sec:challenges}. For many people, the ultimate goal of capturing the meaning of a sentence in a computer's memory might currently seem rather utopic, something of theoretical only interest for the researchers to play with. That would be wrong; computers are already capable of doing a lot of amazing things: they can adequately translate text from one language to another, they respond to vocal instructions, they score to TOEFL\footnote{Test of English as a Foreign Language.} tests at least as well as human beings do (see, for example, \cite{Landauer}), and---perhaps the most important of all---they offer us all the knowledge of the world from the convenience of our desk with just few mouse clicks. What at least I hope is apparent from the current presentation is that compositional-distributional models of meaning is a technology that slowly but steadily evolves to a \textit{useful tool}, which is after all the ultimate purpose of every scientific research.

\section*{Acknowledgements}

This paper discusses topics that I study and work on the past two and a half years, an endeavour that would be doomed to fail without the guidance and help of Mehrnoosh Sadrzadeh, Stephen Pulman and Bob Coecke. I would also like to thank my colleagues in Oxford Edward Grefenstette and Nal Kalchbrenner for all those insightful and interesting discussions. The contribution of the two anonymous reviewers to the final form of this article was invaluable; their suggestions led to a paper which has been tremendously improved compared to the original version. Last, but not least, support by EPSRC grant EP/F042728/1 is gratefully acknowledged.

\nolinenumbers

\bibliographystyle{apalike}
\bibliography{refs}

\end{document}